\documentclass{sig-alternate}

\usepackage{subfig}
\usepackage{graphicx}
\usepackage{color}
\usepackage{epstopdf}
\usepackage[turnoff]{notes}
\usepackage{algorithm}
\usepackage{algorithmic}

\makeatletter
\def\@copyrightspace{\relax}
\makeatother

\newcommand{\vect}[1]{\boldsymbol{#1}}

\title{Factorizing LambdaMART for cold start recommendations}

\numberofauthors{3}
\author{
	\alignauthor Phong Nguyen\\
       \affaddr{AI Group, CUI, CS Dept}\\
       \affaddr{University of Geneva}\\
       \affaddr{Geneva, Switzerland}\\
       \email{Phong.Nguyen@unige.ch}
       \alignauthor Jun Wang\\
       \affaddr{Expedia, Inc}\\
       \affaddr{12, Rue du Lac}\\
       \affaddr{Geneva, Switzerland}\\
       \email{jwang1@expedia.com}
       \alignauthor Alexandros Kalousis\\
       \affaddr{Business Informatics Dept}\\
       \affaddr{University of Applied Sciences}\\
       \affaddr{Western Switzerland}\\
       \email{Alexandros.Kalousis@hesge.ch}
}

\begin{document}
\maketitle

\begin{abstract}
Recommendation systems often rely on point-wise loss metrics such as the mean squared error. However, in real recommendation settings 
only few items are presented to a user. This observation has recently encouraged the use of rank-based metrics. LambdaMART is the 
state-of-the-art algorithm in learning to rank which relies on such a metric. Despite its success it does not have a principled
regularization mechanism relying in empirical approaches to control model complexity leaving it thus prone to overfitting.

Motivated by the fact that very often the users' and items' descriptions as well as the preference behavior can be well 
summarized by a small number of hidden factors, we propose a novel algorithm, LambdaMART Matrix Factorization 
(LambdaMART-MF), that  learns a low rank latent representation of users and items using gradient boosted trees.
The algorithm factorizes lambdaMART by defining relevance scores as the inner product of the 
learned representations of the users and items. The low rank is essentially a model complexity controller; 
on top of it we propose additional regularizers to constraint the learned latent representations that reflect
the user and item manifolds as these are defined by their original feature based descriptors and the
preference behavior. Finally we also propose to use a weighted variant of NDCG to reduce the penalty for 
similar items with large rating discrepancy.

We experiment on two very different recommendation datasets, meta-mining and movies-users, 
and evaluate the performance of LambdaMART-MF, with and without regularization, in the cold 
start setting as well as in the simpler matrix completion setting. In both cases 
it outperforms in a significant manner current state of the art algorithms.
%
\end{abstract}

\category{H.4}{Information Systems Applications}{cold start, recommender systems, learning to rank}



\section{Introduction}

\note[basic idea]{
Recommendation originally focus on MSE error. However, many application in recommendation require the learning to rank idea.
the basic idea of learning to rank to recommendation. It has been explored in some paper..
In this paper, we propose to introduce lambdaMART for recommendation. However, it is easy to overfit with no regularization in its formulation.
To address this limitation, I propose to factorized lambdaMART for recommendation....}

Most recommendation algorithms minimize a point-wise loss function such as the mean squared error or the mean average error 
between the predicted and the true user preferences. For instance in matrix factorization, a learning paradigm very popular
in recommendation problems, state-of-the-art approaches such as~\cite{Srebro2005,Abernethy2006,Agarwal2010} minimize the 
squared error between the inner product of the learned low-rank representations of users and items and the respective 
true preference scores. Such cost functions are clearly not appropriate for recommendation problems 
since what matters there is the rank order of the preference scores and not their absolute values, i.e.
items that are very often top ranked should be highly recommended. It is only recently that recommendation 
methods have started using ranking-based loss functions in their 
optimization problems. Cofirank~\cite{weimer2007maximum}, a collaborative ranking algorithm,  
does maximum-margin matrix factorization by optimizing an upper bound of NDCG measure, a ranking-based
loss function. However, like many recommendation algorithms, it cannot address the cold start problem, 
i.e. it cannot recommend new items to new users.

In preference learning~\cite{furnkranz2010preference} we learn the preference order of documents for a given query. Preference learning algorithm 
are used extensively in search engines and IR systems and optimize ranking-based loss functions such as NDCG. Probably the best known example of 
such algorithms is LambdaMART, \cite{burges2010ranknet}, the state-of-the-art learning to rank algorithm. Its success is due to its 
ability to model preference orders using features describing side-information of the query-document pairs in a very flexible manner.  
Nevertheless it is prone to overfitting because it does not come with a rigorous regularization formalization and relies instead 
on rather empirical approaches to protect against it, such as validation error, early stopping, restricting the size of the base 
level trees etc. \note[Alexandros-Jun2014]{What are the ways 
in which LambdaMart controls overfitting?  I have given two, validation error, early stopping, are they correct? if not then correct/complete.}
\note[Phong-Jun2014]{These are the two most important. There is also the shrinkage parameter which controls the learning rate but this is common for any gradient descent algorithms, so I think we dont need to mention it.}

In this paper we develop a new recommendation algorithm with an emphasis on the cold start problem and the exploitation of available
side-information on users and items. Our algorithm can be thought of as a variant of LambdaMART in which instead of directly learning the user-item preferences, 
as LambdaMART does, we first learn low rank latent factors that describe the users and the items and use them to compute the user-item 
preference scores; essentially our algorithm does a low-rank matrix factorization of the preferences matrix. Moreover we provide additional 
data-based regularizers for the learned latent representations of the users and items that reflect the user and item manifolds as they are 
established from their side-information  as well as from the preference matrix. We evaluate the performance of our algorithm on two very
different recommendation applications and compare its performance to a number of baselines, amongst which LambdaMART, and demonstrate very significant performance
improvements.

\section{Preliminaries}
\label{seq:prel}
We are given a (sparse) preference matrix, ${\mathbf Y}$, of size $n \times m$. The $(i,j)$ non-missing entry of ${\mathbf Y}$
represents the preference score of the $i$th user for the $j$th item in recommendation problems or the relevance score
of the $j$th document for the $i$th query in learning to rank problems, the larger the value of the $(i,j)$ entry the larger
the relevance or preference is. \note[Alexandros-June2014]{Does a large value of $y_{ij}$ a large preference?}
\note[Phong-Jun2014]{Yes. And conversely we order the preference values decreasingly so that a low rank means a high preference.}
In addition to the preference matrix, we also have the descriptions of the users and items. 
We denote by $\vect c_i = (c_{i1},\dots,c_{id})^\text{T} \in \mathbb R^{d}$, the $d$-dimensional description of the $i$th user, and by $\mathbf C$ 
the $n \times d$ user description matrix, the $i${th} row of which is given by the $\vect c_i^T$. Similarly, we denote 
by $\vect d_j = (d_{j1},\dots,d_{jl})^\text{T} \in \mathbb R^{l}$, the $l$-dimensional description of the $j$th item and by $\mathbf D$ 
the $m \times l$ item description matrix, the $j${th} row of which is given by the $\mathbf d_j^T$.

As already mentioned when recommending an item to a user we only care about the rank order of $y_{ij}$ and not the actual preference
score value $y_{ij}$. Thus in principle a preference learning algorithm does not need to predict the 
exact value of $y_{ij}$ but only its rank order as this induced by the preference vector $\vect y$. 
We will denote by $\vect r_{\vect c_i}$ the target rank vector of the $i$th user. We construct $\vect r_{\vect c_i}$ by ordering in a decreasing manner the $i$th 
user's non-missing preference scores over the items; its $k$th entry, $\vect r_{{\vect c_i}k}$, is the rank of the $k$th 
non-missing preference score, with the highest preference score having a rank of one. 
\note[Alexandros-June2014]{How is the rank done increasing or decreasing? related to that, a large preference leads to high rank, i.e. a large
value of $\vect r_{{\vect c_i}k}$ or a low value of $\vect r_{{\vect c_i}k}$, in other words the top ranked item has $\vect r_{{\vect c_i}k}=1$
or $\vect r_{{\vect c_i}k}=m$ where $m$ is the number of items.}
\note[Phong-Jun2014]{See above.}
In the inverse problem, i.e. matching users to items, we will denote by $\vect r_{\vect d_j}$ the target rank vector of the 
$j$th item, given by ordering the $j$th item's non-missing relevance scores for the users.

\subsection{Evaluation Metric}
In real applications of preference learning, such as recommendation, information retrieval, etc, only a few top-ranked items are finally
shown to the users. As a result appropriate evaluation measures for preference learning focus on the correctness of the top-ranked
items. One such, very often used, metric is the {\em Discounted 
Cumulative Gain} (DCG)\cite{jarvelin2000ir}, which is defined as follows:
\begin{eqnarray}
\label{dcg-definition}
\text{DCG}(\vect r,\vect y)@k &=& \sum_{i=1}^M \frac{2^{y_i}-1}{\log_2(r_ i+1 )} I(r_i \le k) 
\end{eqnarray}
\note[Alexandros-June2014]{Please check the formula, I have changed $ I(r_i \ge k)$ to $ I(r_i \le k)$ because the former does not seem
to be correct. To my understaing the most relevant documents, i.e. high $y_{i}$ get a small value of $r_i$, so if you emphasize
on the top ranked positions then necessarily $r_i$ should be small.}
\note[Phong-Jun2014]{Exact. See above.}
where $k$ is the truncation level at which DCG is computed and  $I$ is the indicator function which returns $1$ if its argument holds otherwise $0$. 
$\vect y$ is the $m$ dimensional ground truth relevance vector and $\vect r$ is a rank vector that we will learn. The DCG score measures the match between the
given rank vector $\vect r$ and the rank vector of the relevance score vector $\vect y$. It is easy to check that 
if the rank vector $\vect r$ correctly preserves the order induced by $\vect y$ then the DCG score will achieve its maximum. 
Due to the log in the denominator the DCG score will incur larger penalties when misplacing top items compared to than low end 
items, emphasizing like that the correctness of top items. 

Since the DCG score also depends on the length of the relevance vector $\vect y$, it is often normalized with respect to its maximum score, resulting to 
what is known as the Normalized DCG (NDCG), defined as:
\begin{eqnarray}
\label{ndcg-definition}
\text{NDCG}( \vect y, \hat{\vect y})@k &=& \frac{\text {DCG}(r(\hat{\vect y}),\vect y)@k}{\text {DCG}(r(\vect y),\vect y)@k} 
\end{eqnarray}
$r(\cdot)$ is a rank function that outputs the rank vector, in decreasing order, of its input argument vector.
Thus the vector $r({\vect y})$ is the rank vector of 
the ground truth relevance vector $\vect y$ and $r(\hat{\vect y})$ is the rank vector of the predicted relevance 
vector $\hat{\vect y}$ provided by the learned model. With normalization, 
the value of NDCG ranges from $0$ to $1$, the larger the better. In this paper, we will also use this metric as our evaluation metric.

\subsection{LambdaMART}
The main difficulty in learning preferences is that rank functions are not continuous and have combinatorial complexity. 
Thus most often instead of the rank of the preference scores the pairwise order constraints over the items' preferences are used.
LambdaMART is one of the most popular algorithms for preference learning which follows exactly this 
idea,~\cite{burges2010ranknet}. Its optimization problem relies on a distance
distribution measure, cross entropy, between a learned distribution\footnote{This
learned distribution is generated by the sigmoid function $P^i_{jk}=\frac{1}{1 + e^{\sigma (\hat{y}_{ij} - \hat{y}_{ik})} }$ of the estimated 
preferences $\hat{y}_{ij} , \hat{y}_{ik}$.} 
that gives the probability that item $j$ is more relevant than item $k$  from the true distribution which has a probability mass 
of one if item $i$ is really more relevant than  item $j$ and zero otherwise. \note[Jun]{This confuse me a lot. Do you really need that detail? It will be never used again.}
The final loss function of LambdaMart defined over all users $i$ and overall the respective pairwise preferences for items $j$, $k$, 
is given by:
\begin{eqnarray}
\label{loss:lambdaMART}
\mathcal L(\mathbf Y, \hat {\mathbf Y})  = \sum^n_{i=1} \sum_{\{jk\} \in Z} | \Delta NDCG^i_{jk} |\log(1+e^{-\sigma(\hat{y}_{ij}-\hat{y}_{ik})})
\end{eqnarray}
where $Z$ is the set of all possible pairwise preference constraints such that in the ground truth relevance vector 
holds $y_{ij} > y_{ik}$, and $\Delta NDCG^i_{jk}$ is given by:
$$\Delta NDCG^i_{jk}=NDCG(r(\vect y_i),r(\hat{\vect y}_i))-NDCG(r(\vect y^{jk}_i),r(\hat{\vect y}_i))$$
where $\vect y^{jk}_i$ is the same as the ground truth relevance vector $\vect y_i$ except that the values 
of $y_{ij}$ and $y_{ik}$ are swapped. This is also equal to the NDCG difference that we get if we swap the 
$\hat{y}_{ij}$, $\hat{y}_{ik}$, estimates. Thus the overall loss function of LambdaMART eq~\ref{loss:lambdaMART} 
is the sum of the logistic losses on all pairwise preference constraints weighted by the respective NDCG differences. 
Since the NDCG measure penalizes heavily the error on the top items, the loss function of LambdaMART has also
the same property. LambdaMART minimizes its loss function with respect to all $\hat{y}_{ij}, \hat{y}_{ik}$, and its 
optimization problem is: 
\begin{eqnarray}
\label{opt:lambdaMART}
\min_{\hat{\mathbf Y}} \mathcal L (\mathbf Y, \hat {\mathbf Y}) 
\end{eqnarray}
\cite{yue2007using} have shown empiricially that solving this problem also optimizes the NDCG metric of the 
learned model.
The partial derivative of LambdaMART's loss function with respect to the estimated scores $\hat{y}_{ij}$ is 
\begin{eqnarray}
\label{eq:dlr}
\frac{\partial \mathcal L(\mathbf Y,\hat{\mathbf Y})}{\partial \hat y_{ij}} = \lambda^i_j =  \sum_{\{k|jk\} \in Z} 
\lambda^i_{jk} -  \sum_{\{k|kj\} \in Z} \lambda^i_{kj}
\end{eqnarray} 
and $\lambda^i_{jk}$ is given by:
\begin{eqnarray}
\lambda^i_{jk} = \frac{-\sigma}{1+e^{\sigma(\hat{ y }_{ij} - \hat{ y }_{ik})}} | \bigtriangleup_{NDCG}(\hat{ y }_{ij}, \hat{ y }_{ik}) |
\end{eqnarray}
With a slight abuse of notation below we will write $\frac{\partial \mathcal L(y_{ij},\hat{y_{ij}})}{\partial \hat y_{ij}}$ instead of 
$\frac{\partial \mathcal L(\mathbf Y,\hat{\mathbf Y})}{\partial \hat y_{ij}}$, to make explicit the dependence of the partial derivative
only on $y_{ij},\hat{y}_{ij}$ due to the linearity of $\mathcal L(\mathbf Y, \hat {\mathbf Y})$.

\note[Jun]{This way you first present the gradient descent is also confusing and not natural. I think you should first start from equation \ref{relevance-form}}

LambdaMART uses Multiple Additive Regression Trees (MART)~\cite{friedman2001greedy} to solve its optimization 
problem. It does so through a gradient descent in the functional space that generates preference scores from
item and user descriptions, i.e.
$\hat{y}_{ij}=f(\vect c_i,\vect d_j)$,
where the update of the preference scores at the $t$ step of the gradient descent is given by:
\begin{eqnarray} 
\label{eq:lambda-update-1}
\hat{y}^{(t)}_{ij} = \hat {y}^{(t-1)}_{ij} -  \eta  \frac{\partial \mathcal L(y_{ij}, \hat{y}^{(t-1)}_{ij})}{ \partial \hat{y}^{(t-1)}_{ij} } 
\end{eqnarray}
or equivalently:
\begin{eqnarray} 
\label{eq:lambda-update-2}
f^{(t)}(\vect c_i, \vect d_j) = f^{(t-1)}(\vect c_i, \vect d_j ) - \eta  \frac{\partial \mathcal L(y_{ij}, f^{(t-1)}(\vect c_i, \vect d_j))}
                                                                          {\partial  f^{(t-1)}(\vect c_i, \vect d_j)}
\end{eqnarray}
where $\eta$ is the learning rate.
We terminate the gradient descent when we reach a given number of iterations $T$ or when the validation 
 loss NDCG starts to increase.
We approximate the derivative  $ \frac{\partial \mathcal L(y_{ij}, \hat{y}^{(t-1)}_{ij})}{ \partial \hat{y}^{(t-1)}_{ij} }$ by
learning at each step $t$ a regression tree $h^{(t)}(\vect c, \vect d)$ that fits it by minimizing the sum of squared errors.
Thus at each update step we have 
\begin{eqnarray}
\label{eq:lambda-functional-up-1}
f^{(t)}(\vect c_i, \vect d_j) = f^{(t-1)}(\vect c_i, \vect d_j ) + \eta h^{(t)}(\vect c_i, \vect d_j)
\end{eqnarray}
which if we denote by $\gamma_{tk}$ the prediction of the $k$th terminal node of the $h^{(t)}$ tree 
and by $h_{tk}$ the respective partition of the input space, we can rewrite as:
\begin{eqnarray}
\label{eq:lambda-functional-up-2}
f^{(t)}(\vect c_i,\vect d_j)  = f^{(t-1)}(\vect c_i, \vect d_j ) +  \eta \gamma_{tk} I((\vect c,\vect d) \in h_{tk})
\end{eqnarray}
we can further optimize over the $\gamma_{tk}$ values to minimize the loss function of eq~\ref{loss:lambdaMART} over the instances of each $h_{tk}$ partition 
using Newton's approximation.
The final preference estimation function is given by:
\begin{eqnarray}
\label{relevance-form-a}
\hat{y}=f(\vect c,\vect d) = \sum_{t=1}^T \eta h^{(t)}(\vect c, \vect d)
\end{eqnarray}
or
\begin{eqnarray}
\label{relevance-form}
\hat{y}=f(\vect c,\vect d)=\sum^T_{t=1}\eta \gamma_{tk} I((\vect c,\vect d) \in h_{tk})
\end{eqnarray}
\note[Alexandros]{This is taken from the review paper~\cite{burges2010ranknet}, it seems to be saying that we specifically 
optimize each $\gamma_{tk}$ using Newton's method. It is not clear to me how this is done, because it seems to assume that
the tree structure is given. It does not seem to be in aggreement with the standard way of growing a regression tree, i.e.
choose the test that at each node it minimizes the sum of squares error. Moreover when I asked Phong he pointed me to line
search. There is something I am missing here.}
\note[Alexandros]{
Ok after discussing with Phong it is a bit more clear now. I am not sure we need all the blabla above. It should be enough
to keep one of the pairs of Eqs~\ref{eq:lambda-functional-up-1}~\ref{relevance-form-a}, 
or~\ref{eq:lambda-functional-up-2}~\ref{relevance-form}.
}

 

LambdaMart is a very effective algorithm for learning to rank problems, see e.g \cite{burges2011learning,donmez2009local}. It learns 
non-linear relevance scores, $\hat{y}_{ij}$, using gradient boosted regression trees. The number of the 
parameters it fits is given by the number of available preference scores (this is typically some fraction of $n \times m$); 
there is no regularization on them to prevent overfitting.  The only protection against overfitting can come from rather 
empirical approaches such as constraining the size of the regression trees or by selecting learning rate $\eta$.

\section{Factorized Lambda-MART}
\label{sec:lmmf}
In order to address the rather ad-hoc approach of LambdaMART to prevent overfitting we propose here a factorized variant of it that does regularization in a principled manner. 
Motivated by the fact that very often the users' and the items' descriptions and their preference relations can be well summarized by a small number of hidden factors we learn 
a low rank hidden representation of users and items using gradient boosted trees, MART. We define the relevance score as the inner product of the new representation of the users
and items which has the additional advantage of  introducing low rank structural regularization in the learned preference matrix.

Concretely, we define the relevance score of the $i$th user and $j$th item by  $\hat y_{ij}=\vect u_i^\text{T} \vect v_j$, 
where $\mathbf u_i$ and $\mathbf v_j$ are the $r$-dimensional user and item latent descriptors. We denote by 
$\mathbf U: n \times r $ and $\mathbf V: m \times r $ the new representation matrices of users and items. The dimensionality of $r$ is a small 
number, $r << \min(n,m)$. 
The loss function of eq~\ref{loss:lambdaMART} now becomes:
{\small
\begin{eqnarray}
\label{eq:lrmf}
\mathcal L_{MF}(\mathbf Y, \hat{ \mathbf U}, \hat{ \mathbf V}) = 
\sum_{i=1}^n \sum_{\{jk\} \in Z}  | \bigtriangleup_{NDCG} | \log(1+e^{-\sigma(\hat{\vect u}_i (\hat{\vect v}_j - \hat{\vect v}_k) )})
\end{eqnarray} }
The partial derivatives of this lost function with respect to $\hat{\vect u}_i, \hat{\vect v}_j$, are given by:
\begin{eqnarray}
\frac{\partial \mathcal L_{MF}(\mathbf Y, \hat{\mathbf U}, \hat{\mathbf V})}
     {\partial \hat{\vect u}_i} =  
\sum_{j=1}^m \frac{\partial \mathcal L_{MF}(\mathbf Y, \hat{\mathbf Y})}
                  {\partial \hat{y}_{ij}}
             \frac{\partial \hat{y}_{ij}}
                  {\partial \hat{\vect u}_i} =
\sum_{j=1}^m \lambda^i_j 
             \frac{\partial \hat{y}_{ij}}
                  {\partial \hat{\vect u}_i} \label{eq:pdu}  \\
\frac{\partial \mathcal L_{MF}(\mathbf Y, \hat{\mathbf U}, \hat{\mathbf V})}
     {\partial \hat{\vect v_j}} =  
\sum_{i=1}^n \frac{\partial \mathcal L_{MF}(\mathbf Y, \hat{\mathbf Y})}
                  {\partial \hat{y}_{ij}}
             \frac{\partial \hat{y}_{ij}}
                  {\partial \hat{\vect v}_i} =
\sum_{i=1}^n \lambda^i_j 
              \frac{\partial \hat{y}_{ij}}
                   {\partial \vect v_j} \label{eq:pdv}
\end{eqnarray}
Note that the formulation we give in equation \ref{eq:lrmf} is very similar to those used
in matrix factorization algorithms. Existing matrix factorization algorithms used in collaborative
filtering recommendation learn the low-rank representation of users and items in order to complete the sparse 
preference matrix $\mathbf Y$, \cite{Srebro2005,weimer2007maximum,Abernethy2006,Agarwal2010}, however these
approaches cannot address the cold start problem. 


Similar to LambdaMART we will seek a function $f$ that will optimize the $\mathcal L_{MF}$ 
loss function. To do so we will learn functions of the latent profiles of the users and items
from their side information. We will factorize $f(\mathbf c, \mathbf d)$ by  
$$f(\mathbf c, \mathbf d) = f_u(\mathbf c)^\text{T}f_v(\mathbf d) = \mathbf u^\text{T}\mathbf v = \hat y$$
where $f_u: \mathcal C \rightarrow \mathcal U$ is a learned user function that gives us
the latent factor representation of user from his/her side information descriptor $\mathbf c$;
$f_v: \mathcal D \rightarrow \mathcal V$ is the respective function for the items.
\note[Alexandros]{
Note that this is a similar formulation to functional matrix factorization \cite{Zhou2011}. 
Does this mean that this guys have done the same thing, using instead of the
ensemble trees a single tree? If yes that can create a problem on the novelty of the paper.
Anyway this should now go on a related work section.}
We will follow the LambdaMART approach
described previously and learn an ensemble of trees for each one of the  $f_u$ and $f_v$ functions.
Concretely:
\begin{eqnarray}
\hat{\mathbf u}_i &=& f_u(\mathbf c_i)  = \sum_{t=1}^T \eta h_{u}^{(t)}(\mathbf c_i)  \label{eq:u-function} \\
\hat{\mathbf v}_j &=& f_v(\mathbf d_j)  = \sum_{t=1}^T \eta h_{v}^{(t)}(\mathbf d_j)  \label{eq:v-function}  
\end{eqnarray}
Unlike standard LambdaMART the trees we will learn are multi-output regression trees, predicting the
complete latent profile of users or items. 
\note[Alexandros]{And how are these trees grown? What kind of measure we optimise?  Is the
description I give enough, or is there something more to add?}
Now at each step $t$ of the gradient descent we
learn the $h^{(t)}_{u}(\mathbf c)$ and $h_{v}^{(t)}(\mathbf d)$ trees 
that fit the negative of the partial derivatives given at equations~\ref{eq:pdu},~\ref{eq:pdv}. 
We learn the trees by greedily optimizing the sum of squared errors over the over the dimensions
of the partial gradients. 
The $t$ gradient descent step for $\vect u_i$ is given by:
$$\vect u_i^{(t)}  = \mathbf u_i^{(t-1)} - \sum_{j=1}^m 
             \frac{\partial \mathcal L_{MF}(\mathbf Y, \hat{\mathbf Y})}
                  {\partial \hat{y}_{ij}} \Big|_{\hat{y}_{ij}=\hat{\vect u}^{(t-1)\text{T}}_{i} {\hat{\vect v}_j^{(t-1)}}}
             \frac{\partial \hat{\vect u}_{i}^\text{T} {\hat{\vect v}_j^{(t-1)}}}
                  {\partial \hat{\vect u}_i} $$
%
and for $\vect v_j$ by:
$$\vect v_j^{(t)}  = \vect v_j^{(t-1)} - \sum_{i=1}^n 
             \frac{\partial \mathcal L_{MF}(\mathbf Y, \hat{\mathbf Y})}
                  {\partial \hat{y}_{ij}} \Big|_{\hat{y}_{ij} = \hat{\vect u}_{i}^{(t-1)\text{T}} {\hat{\vect v}_j^{(t-1)}}}
             \frac{\partial \hat{\vect u}_{i}^{(t-1)\text{T}} {\hat{\vect v}_j}}
                  {\partial \hat{\vect v}_j} $$
The functions of each step are now:
\begin{eqnarray}
f_{\vect u}^{(t)}(\vect c) = f_{u}^{(t-1)} (\vect c) + \eta h^{(t)}_{u} (\vect c) \\
f_{\vect v}^{(t)}(\vect d) = f_{v}^{(t-1)} (\vect d) + \eta h^{(t)}_{v} (\vect d)
\end{eqnarray}
which give rise to the final form functional estimates that we already gave in equations~\ref{eq:u-function} and~\ref{eq:v-function} 
respectively. Optimizing for both  $\vect u_i$ and $\vect v_j$ at each step of the gradient descent 
results to a faster convergence, than first optimizing for one while keeping the other fixed and vice versa.
We will call the resulting algorithm LambdaMART Matrix Factorization and denote it by LM-MF.
\note[Alexandros]{What are the termination criteria here?}

\section{Regularization}
\label{sec:reg}

In this section, we will describe a number of different regularization methods to constraint in a meaningful manner 
the learning of the user and item latent profiles. We will do so by incorporating different regularizers inside the 
gradient boosting tree algorithm to avoid overfitting during the learning of the $f_u$ and $f_v$ functions. 

\subsection{Input-Output Space Regularization}

LambdaMART-MF learns a new representation of users and items. We will regularize these representations by constraining
them by the geometry of the user and item spaces as this is given by the $\mathbf c$ and $\mathbf d$ descriptors respectively.
Based on these descriptors we will define user similarities and item similarities, which we will call input-space similarities
in order to make explicit that they are compute on the side-information vectors describing the users and the items. In addition
to the input-space similarities we will also define what we will call output space similarity which will reflect the
similarities of users(items) according to the respective similarities of their preference vectors. We will regularize the
learned latent representations by the input/output space similarities constraining the former to follow the latter.

To define the input space similarities we will use  the descriptors of users and items. Concretely given two users 
$\vect c_i$ and $\vect c_j$ we measure their input space similarity $s_{U_{in}}(\vect c_i, \vect c_j)$   
using the heat kernel over their descriptors as follows: 
\begin{eqnarray}
s_{U_{in}}(\vect c_i, \vect c_j; \sigma) &=& e^{-\sigma || \vect c_i - \vect c_j ||^2} 
\end{eqnarray}
where $\sigma$ is the inverse kernel width of the heat kernel;  we set its value to the squared inverse average Euclidean distance of all 
the users in the $\mathcal C$ space, i.e. $\sigma = (( \frac{1}{n} \sum_{ij} || \vect c_i - \vect c_j ||  )^2)^{-1}$.  
By applying the above equation over all user pairs, we get the $\mathbf S_{U_{in}}: n \times n$ user input similarity matrix. 
We will do exactly the same to compute the $\mathbf S_{V_{in}}: m \times m$ item input similarity matrix using the item descriptors $\mathbf D$. 

To define the output space similarities we will use  the preference vectors of users and items. 
Given two users $i$ and $j$ and their preference vectors $\vect y_{i\cdot}$ and $\vect y_{j\cdot}$, 
we will measure their output space similarity $s_{U_{out}}(\mathbf y_{i\cdot}, \mathbf y_{j\cdot})$ 
using NDCG@k since this is the metric that we want to optimize. 
To define the similarities, we first compute the NDCG@k on the  $(\vect y_{i\cdot}, \vect y_{j\cdot})$ pair
as well as on the  $(\vect y_{j\cdot}, \vect y_{i\cdot})$, because the NDCG@k is not symmetric,  and compute
the distance $d_{\text {NDCG}@k}(\mathbf y_{i\cdot}, \mathbf y_{j\cdot})$ between the two preference vectors as the average 
of the  two previous NDCG@k measures which we have subtracted from one:
\begin{eqnarray}
d_{\text {NDCG}@k}(\mathbf y_{i\cdot}, \mathbf y_{j\cdot}) &=& \frac{1}{2}((1- \text {NDCG}@k(\mathbf y_{i\cdot}, \mathbf y_{j\cdot})) \nonumber \\
& &+ (1-\text {NDCG}@k(\mathbf y_{j\cdot}, \mathbf y_{i\cdot}))   ) \nonumber 
\end{eqnarray} 
We define  finally the output space similarity $s_{U_{out}}(\mathbf y_{i\cdot}, \mathbf y_{j\cdot})$ by the  exponential of the negative  distance: 
\begin{eqnarray}
\label{eq:outsim}
s_{U_{out}}(\mathbf y_{i\cdot}, \mathbf y_{j\cdot}) &=& e^{- d_{\text {NDCG}@k}(\mathbf y_{i\cdot}, \mathbf y_{j\cdot})} 
\end{eqnarray}
The resulting similarity measure gives high similarity to preference vectors that are very similar in their 
top-$k$ elements, while preference vectors which are less similar in their top-$k$ elements will get much lower similarities.  
We apply this measure over all user preference vector pairs to get the $\mathbf S_{U_{out}}: n \times n$ user output similarity matrix. 
We do the same for items using now the $\mathbf y_{\cdot i}$ and $\mathbf y_{\cdot j}$ preference vectors for each $ij^{th}$ item 
pair to get the $\mathbf S_{V_{out}}: m \times m$ item output similarity matrix.

To regularize the user and item latent profiles, we will use graph laplacian regularization and force them 
to reflect the manifold structure of the users and items as these are given by the input and outpout space 
similarity matrices. 
Concretely, we define the user and item regularizers $\mathcal R_U$ and $\mathcal R_V$ as follows: 
\begin{eqnarray}
\mathcal R_U = \mu_1 || \hat{ \mathbf U}^\text{T}  \mathbf  L_{U_{in}} \hat{ \mathbf U} ||^2_F + \mu_2 || \hat{ \mathbf U}^\text{T}  \mathbf  L_{U_{out}} \hat{ \mathbf U} ||^2_F \\
\mathcal R_V = \mu_1 || \hat{ \mathbf V}^\text{T}  \mathbf  L_{V_{in}} \hat{ \mathbf V} ||^2_F + \mu_2 || \hat{ \mathbf V}^\text{T}  \mathbf  L_{V_{out}} \hat{ \mathbf V} ||^2_F
\end{eqnarray}
where the four laplacian matrices $\mathbf L_{U_{in}}$, $\mathbf L_{U_{out}}$, $\mathbf L_{V_{in}}$ and $\mathbf L_{V_{out}}$ are 
defined as $\mathbf L = \mathbf D - \mathbf S$ where $\mathbf D_{ii} = \sum_{j}  \mathbf S_{ij}$ and $\mathbf S$ are the corresponding
similarity matrices. $\mu_1$ and $\mu_2$ are regularization parameters that control the relative importance of the input and output space 
similarities respectively.

\note[Removed]{We can furthermore justify the use of regularization in the specific MART context as follows. 
Very often, the description of users and items contains irrelevant input features. 
By constraining the new representation of users and items by their input-output similarities, 
users with similar latent profiles will end up into the same terminal node of trees. 
Thus, the gradient boosted trees will be constructed by acknowledging the informative 
features of user descriptions  such that the irrelevant features will be automatically 
removed. This motivation is similar to distance metric learning, which learns a new representation 
of learning instances in which learning instances with same class will have smaller distances. 
Distance learning  has been shown to be robust to irrelevant features~\cite{weinberger2009distance,wang2012parametric}.}
\note[Alexandros-June2014]{The above paragraph, as it is written does not say anything, nor
does it bring any information, so I jsut remove it. If you thin there is an important message
to pass you should explain it more clear.}
\note[ToCheck]{I am not sure I get the meaning of the above paragraph}
\note[Phong-June2014]{Well, this paragraph was from Jun who tried to link the algorithm with metric learning. I agree that we can remove it.}

\subsection{Weighted NDCG Cost}
\label{sec:wndcg}

In addition to the graph laplacian regularization over the latent profiles, we also provide a soft variant of the NDCG loss used in LambdaMART. 
Recall that in NDCG the loss is determined by the pairwise difference incurred if we exchange the position of two items $j$ and $k$ for 
a given user $i$.  This loss can be unreasonably large even if the two items are similar to each other with respect to the similarity measures defined above. 
A consequence of such large penalties will be a large deviance of the gradient boosted trees under which similar items will not anymore fall in the same leaf 
node.  To alleviate that problem we introduce a weighted NDCG difference which takes into account the items' input and output similarities, which we define
as follows: 
\begin{eqnarray}
\mathbf S_V &=& \mu_1 \mathbf S_{V_{in}} + \mu_2 \mathbf S_{V_{out}}  \\
\bigtriangleup {WNDCG}^i_{jk} &=&  \bigtriangleup {NDCG}^i_{jk}  (1-s_{V_{jk}})
\end{eqnarray}
Under the weighted variant if two items $j$ and $k$ are very similar the incurred loss will be by construction 
very low leading to a smaller loss and thus less deviance of the gradient boosted trees for the two items.

\subsection{Regularized LambdaMART-MF}
By combing the input-output space regularization and the weighted NDCG with LM-MF  given in equation \ref{eq:lrmf} we obtain the regularised LambdaMART matrix factorization
the objective function of which is
\begin{eqnarray}
\label{eq:lrmf_all}
\mathcal L_{RMF}(\mathbf Y, \hat{ \mathbf U}, \hat{ \mathbf V}) = \nonumber \\ 
\sum_{i=1}^n \sum_{\{jk\} \in Z}  | \bigtriangleup_{WNDCG} | \log(1+e^{-\sigma(\hat{\vect u}_i (\hat{\vect v}_j - \hat{\vect v}_k) )}) + \mathcal R_U + \mathcal R_V  \nonumber \\ 
\end{eqnarray}
Its partial derivatives with respect to $\hat{\vect u}_i, \hat{\vect v}_j$, are now given by:
\begin{eqnarray}
\label{eq:partialu}
\frac{\partial \mathcal L_{RMF}(\mathbf Y, \hat{\mathbf U}, \hat{\mathbf V})}
     {\partial \hat{\vect u}_i} &=&  
\sum_{j=1}^m \lambda^i_j 
             \frac{\partial \hat{y}_{ij}}
                  {\partial \hat{\vect u}_i}  \nonumber \\
& &+ 2 \mu_1 \sum_{j \in \mathbb N^i_{U_{in}}} s_{U_{in}, {ij}} (\hat{\vect u}_i - \hat{\vect u}_j ) \nonumber \\
& &+ 2 \mu_2 \sum_{j  \in \mathbb N^i_{U_{out}}} s_{U_{out}, {ij}} (\hat{\vect u}_i - \hat{\vect u}_j ) \label{eq:pdu_all}  
\end{eqnarray}
\begin{eqnarray}
\label{eq:partialv}
\frac{\partial \mathcal L_{RMF}(\mathbf Y, \hat{\mathbf U}, \hat{\mathbf V})}
     {\partial \hat{\vect v_j}} &=&  
\sum_{i=1}^n \lambda^i_j 
              \frac{\partial \hat{y}_{ij}}
                   {\partial \vect v_j} \nonumber \\
& &+ 2 \mu_1 \sum_{i \in \mathbb N^j_{V_{in}}} s_{V_{in}, {ij}} (\hat{\vect v}_j - \hat{\vect v}_i ) \nonumber \\
& &+ 2 \mu_2 \sum_{i \in \mathbb N^j_{V_{out}}} s_{V_{out}, {ij}} (\hat{\vect v}_j - \hat{\vect v}_i ) \label{eq:pdv_all}
\end{eqnarray}
where $\mathbb N^i_{U_{in}}$ is the set of the $k$ nearest neighbors of the $i$th user defined on the basis of the input similarity.

To optimize our final objective function, equation \ref{eq:lrmf_all}, we learn the latent profiles of users and items by gradient 
boosted trees. We will call the resulting algorithm Regularized LambdaMART Matrix Factorization and denote it by LM-MF-Reg. The 
algorithm is described in Algorithm \ref{algo:LM-MF-Reg}. At $t$ iteration, we first compute the partial derivatives of the objective 
function at point $(\hat{\mathbf U}^{t-1},\hat{\mathbf V}^{t-1})$. Then we fit the trees $h^t_{\vect u}$ and $h^t_{\vect v}$ 
for the user and item descriptions respectively. Finally, we update the predictions of $\hat{\mathbf U}$ and $\hat{\mathbf V}$ according to the output 
of regression trees. The learning process is continued until the maximum number of trees is reached or the early stop criterion is satisfied. In all 
our experiments, the maximum number of trees is set by 15000. The early stopping criterion is no validation set error improvement in 200 iterations. 

\begin{algorithm}[t]
   \caption{Regularized LambdaMART Matrix Factorization}
   \label{algo:LM-MF-Reg}
   \begin{algorithmic}
   \STATE {\bfseries Input:} $\mathbf{C}$, $\mathbf{D}$,$\mathbf{Y}$, $\mathbf{S_{\mathbf U_{out}}}$,$\mathbf{S_{\mathbf U_{in}}}$, $\mathbf{S_{\mathbf V_{out}}}$,$\mathbf{S_{\mathbf V_{in}}}$,$\mu_1$, $\mu_2$, $\eta$,$r$, and $T$
   \STATE {\bfseries Output:} $f_{\vect u}$ and $f_{\vect v}$
   
   \STATE initialize: $f^0_{\vect u}(\vect c)$ and $f^0_{\vect v}(\vect d)$ with random values 
	 \STATE initialize $t=1$
	 \REPEAT
   \STATE a) compute $\vect r^t_{{\vect u_i}}=-\frac{\partial \mathcal L_{RMF}(\mathbf Y, \hat{\mathbf U}, \hat{\mathbf V})} {\partial \hat{\vect u}_i}\left|_{\hat{\mathbf U}=\hat{\mathbf U}^{t-1},\hat{\mathbf V}=\hat{\mathbf V}^{t-1}} \right.$ according to equation \ref{eq:partialu}, for $i= 1 $ to $ n$
	\STATE b) compute $\vect r^t_{{\vect v_j}}=-\frac{\partial \mathcal L_{RMF}(\mathbf Y, \hat{\mathbf U}, \hat{\mathbf V})} {\partial \hat{\vect v_j}}\left|_{\hat{\mathbf U}=\hat{\mathbf U}^{t-1},\hat{\mathbf V}=\hat{\mathbf V}^{t-1}} \right.$ according to equation \ref{eq:partialv},  for $j= 1 $ to $ m$
	\STATE c)fit a multi-output regression tree $h^t_{\vect u}$ for $\{(\vect c_i,\vect r^t_{\vect u_i})\}^n_{i=1}$
	\STATE d)fit a multi-output regression tree $h^t_{\vect v}$ for $\{(\vect d_i,\vect r^t_{\vect v_i})\}^m_{i=1}$
	\STATE e) $f^t_{\vect u}(\vect c_i) = f^{t-1}_{\vect u}(\vect c_i)+\eta h^t_{\vect u}(\vect c_i)$
	\STATE f) $f^t_{\vect v}(\vect d_j) = f^{t-1}_{\vect v}(\vect d_j)+\eta h^t_{\vect v}(\vect d_j)$
   \UNTIL {converges or t=T}
\end{algorithmic}
\end{algorithm} 


\section{Experiments}

We will evaluate the two basic algorithms that we presented above, LM-MF and LM-MF-Reg, on two recommendation problems, 
meta-mining and MovieLens, and compare their performance to a number of baselines. 

Meta-mining~\cite{Hilario2010} applies the idea of meta-learning or learning to learn to the whole DM process. 
Data mining workflows and datasets are extensively characterised by side information. The goal 
is to suggest which data mining workflow should be applied on which dataset in view of optimizing some performance measure, e.g. 
accuracy in classification problems, by mining past experiments.  Recently \cite{Nguyen2012} have 
proposed tackling the problem as a hybrid  recommendation problem: dataset-workflow pairs can be seen as user-item 
pairs which are related by the relative performance achieved by the workflows applied on the datasets.  We propose 
here to go one step beyond \cite{Nguyen2012} and consider the meta-mining as a learning to rank problem.  
We should note that meta-mining is a quite difficult problem, because what we are trying to do is in essence 
to predict-recommend what are the workflows that will achieve the top performance on a given dataset.

MovieLens\footnote{http://grouplens.org/datasets/movielens/}  is a well-known benchmark dataset for collaborative filtering. 
It provides one to five star ratings of users for movies, where each user has rated at least twenty movies. In addition the 
dataset provides limited side information on users and items. It has been extensively experimented with most often in 
a non cold start setting, due to the difficulty of exploitation of the available side information. The dataset has two versions: 
the first contains one hundred thousand ratings (100K) and the second one million (1M). We will experiment with both of them.

In Table~\ref{tab:ds} we give a basic description of the three different datasets, namely the numbers of ratings, users, items,
the numbers of user and item descriptors, $d$ and $l$ respectively, and the percentage of available entries in the $\mathbf Y$ 
preference matrix.  The meta-mining dataset has a quite extensive set of features giving the side information and a complete 
preference matrix, it is also considerably smaller than the two MovieLens variants. The MovieLens datasets are characterized
by a large size, the limited availability of side information, and the very small percentage of available entries in the 
preference matrix. Overall we have two recommendation problems with very different characteristics.  

\begin{table}[!t]
\centering
\scalebox{0.8}{
\begin{tabular}{|l|c|c|c|c|c|c|}
\hline
                           & users & items & ratings & \% comp. $\mathbf Y$ & $d$ & $l$ \\ \hline
Meta-Mining                & 65    & 35    & 2275    & 100.0\%              & 113 & 214  \\ \hline
MovieLens 100K             & 943   & 1682  & 100K    &   6.3\%              & 4   & 19 \\ \hline
MovieLens 1M               & 6040  & 3900  & 1M      &   4.2\%              & 4   & 19 \\ \hline
\end{tabular}
}
\caption{Dataset statistics}
\label{tab:ds}
\end{table}

\subsection{Recommendation Tasks}
\label{sec:tasks}
We will evaluate the performance of the algorithms we presented above in two 
different variants of the cold start problem. In the first one that we will call {\em User Cold Start} we will evaluate the
quality of the recommendations they provide to unseen users when the set of items over which we provide recommendations is fixed.
In the second variant, which we will call {\em Full Cold Start}, we will provide suggestions over both unseen users and items.
We will also evaluate the performance of our algorithms in a {\em Matrix Completion} setting; 
here we will randomly remove items from the users preference lists and predict the preferences 
of the removed items with a model learned on the remaining observed ratings. In model-based collaborative 
filtering, matrix completion has been usually addressed by low-rank matrix factorization algorithms that do not exploit side information. 
We want to see whether exploiting the side information can bring substantial improvements in the matrix completion 
performance.  We will do this only for the meta-mining problem because the MovieLens dataset have rather limited side-information. 
\note[Alexandros-June2014]{Have you done the same experiments for movie-lens in the thesis? if not why?}
\note[Phong-June2014]{I did the matrix completion for MovieLens but we were not better than CR. So I let it out of this paper for reason of space, and also I did not include it in the thesis because the thesis is intended for meta-mining, not for MovieLens. }

Note that provision of recommendations in the cold start setting is much more difficult than the matrix completion since in the 
former we do not have historical data for the new users and new items over which we need to provide the recommendations and thus 
we can only rely on their side information to infer the latter.

\subsection{Comparison Baselines}

As first baseline we will use LambdaMART (LM), \cite{burges2010ranknet}, in both cold start variants as well as in the 
matrix completion setting. To ensure a fair 
comparison of our methods against LamdaMART we will train all of them using the same values
over the hyperparameters they share, namely learning rate, size of regression trees and number of iterations.  
In the matrix completion  we will also add CofiRank (CR) as a baseline, a state-of-the-art matrix factorization 
algorithm in collaborative ranking~\cite{weimer2007maximum}. CR minimizes an upper bound of NDCG and uses the $l_2$ norm to regularize the latent factors. 
Its objective function is thus similar to those of our methods and LambdaMART's but it learns directly the latent factors by gradient descent without using the 
side information. 

For the two cold start evaluation variants, we will also have as a second baseline a memory-based approach, one of the most common approaches used for cold start 
recommendations~\cite{bell2007scalable}. In  the  user cold start setting we will provide item recommendations for a new user using its nearest neighbors.
This user memory-based (UB) approach will compute the preference score for the  $j^{th}$ item as:
\begin{eqnarray}
\hat y_j = \frac{1}{|\mathbb N|} \sum_{i \in \mathbb N} y_{ij}
\end{eqnarray}
where $\mathbb N$ is the set of the $5$-nearest neighbors for the new user for which we want to provide the recommendations. We compute 
the nearest neighbors using the Euclidean distance on the user side information.  
In  the  full cold start setting, we will provide recommendations for a new user-item 
pair by joining the nearest neighbors of the new user with the nearest neighbors of the new item. 
This full memory-based (FB) approach will compute the preference score for the  $i^{th}$ user and $j^{th}$ item as:
\begin{eqnarray}
\hat y_{ij} = \frac{1}{|\mathbb N_i| \times |\mathbb N_j|} \sum_{c \in \mathbb N_i} \sum_{d \in \mathbb N_j} y_{cd}
\end{eqnarray}
where $\mathbb N_i$ is the set of the $5$-nearest neighbors for the new user and $\mathbb N_j$ is the set of the $5$-nearest neighbors for the new item.  
Both neighborhoods are computed using the Euclidean distance on the user and item side-information features respectively.  

Finally we will also experiment with the LambdaMART variant in which we use the weighted NDCG cost that we described in 
section~\ref{sec:wndcg} in order to evaluate the potential benefit it brings over the plain NDCG; we will call this method LMW.

\subsection{Meta-Mining}

The meta-mining problem we will consider is the one provided by \cite{Nguyen2012}. It consists of the application of 35 feature selection 
plus classification workflows on 65 real world datasets with genomic microarray or proteomic data related to cancer diagnosis or prognosis, 
mostly from National Center for Biotechnology Information\footnote{http://www.ncbi.nlm.nih.gov/}. 
In \cite{Nguyen2012}, the authors used four feature selection algorithms: Information Gain, {\em IG}, Chi-square, {\em CHI}, 
ReliefF, {\em RF}, and recursive feature elimination with SVM, {\em SVMRFE}; they fixed the number of selected 
features to ten. For classification, they used seven classification algorithms: one-nearest-neighbor,  {\em 1NN}, 
the {\em C4.5} and {\em CART} decision tree algorithms, 
a Naive Bayes algorithm with normal probability estimation, {\em NBN}, a logistic regression algorithm, {\em LR}, and SVM 
with the linear, {\em SVM$_l$} and the rbf, {\em SVM$_r$}, kernels. 
In total, they ended up with $35 \times 65 = 2275$ base-level DM experiments; they used as performance measure the
classification accuracy which they estimated using ten-fold cross-validation. 

The preference score that corresponds to each dataset-workflow pair is given by a number of significance tests 
on the performances of the different workflows that are applied on a given dataset. More concretely, if a workflow is significantly 
better than another one on the given dataset it gets one point, if there is no significant difference between the two workflows
then each gets half a point, and if it is significantly worse it gets zero points. Thus if a workflow outperforms in a statistically
significant manner all other workflows on a given dataset it will get $m-1$ points, where $m$ is the total number of workflows (here 35). 
In the matrix completion and the full cold start settings we compute the preference scores with respect to 
the {\em training} workflows since for these two scenarios the total number of workflows is less than 35.  
In addition, we rescale the performance measure from the 0-34 interval to the 0-5 interval to avoid large
preference scores from overwhelming the NDCG due to the exponential nature of the latter with respect to 
the preference score. 

Finally, to describe the datasets and the data mining workflows we use the same characteristics
that were used in~\cite{Nguyen2012}. Namely 113 dataset characteristics that give statistical, 
information-theoretic, geometrical-topological, landmarking and model-based descriptors of the datasets
and 214 workflow characteristics derived from a propositionalization of a set of 214 tree-structured generalized 
workflow patterns extracted from the ground specifications of DM workflows.


\subsubsection{Evaluation Setting}
We fix the parameters that LambdaMART uses to construct the regression trees to the following values:
the maximum number of nodes for the regression trees is three, the maximum percent of instances in each leaf node is $10\%$ 
and the learning rate, $\eta$, of the gradient boosted tree algorithm to $10^{-2}$.  
To build the ensemble trees of LM-MF and LM-MF-Reg we use the same parameter settings as the ones we use in LambdaMART.  We select their  
input and output regularization parameters $\mu_1$ and $\mu_2$ in the grid $[0.1,1,5,7,10]^2$, by three-fold inner cross-validation.
We fix the number of nearest neighbors for the Laplacian matrices to five. 
To compute the output-space similarities, we used the NDCG similarity measure defined in Eq.~\ref{eq:outsim} 
where the truncation level $k$ is set to the truncation level at which each time we report the results. 

To build the different recommendation scenarios, we have proceeded as follows. 
In matrix completion we randomly select $N$ workflows for each dataset to build the training set. We 
choose $N$ in the range $[5,10,15]$. 
For each dataset, we use ten workflows, different from the $N$ ones selected in training, 
for validation and we use the rest for testing. This scenario emulates a 86\%, 71\% and 57\% of missing values
in the preference matrix. \note[Alexandros-June2014]{In fact this means that your
level of missing values is lower, since the validation set is typically a part of the training set. In
fact you use N in 15, 20 and 25 for training, out of which 10 are kept each time for validation, which 
leads to diffrent levels of missing values. Update the \%.}
\note[Phong-June2014]{No. I randomly select N workflows for training, plus ten for validation and the rest is used for testing.}
Since in the matrix completion setting the numbers of users and items are fixed, we fix the number 
of hidden factors $r$ to $\min(n,m) = 35$ for all three matrix factorization algorithms, LM-MF, LM-MF-Reg and CofiRank. 
For this baseline we used the default parameters as these are provided in \cite{weimer2007maximum}. 
We report the average NDCG@5 measure on the test workflows of each dataset. 

In the user cold start scenario, we will evaluate the performance in providing accurate recommendations
for new datasets. To do so we do a leave-one-dataset-out. We train the different methods on 64 datasets and evaluate 
their recommendations on the left-out dataset. Since there are no missing workflows as previously we also 
fix the number of hidden factors $r$ to $\min(n,m) = 35$. 
In the full cold start scenario on top of the leave-one-dataset-out we also randomly partition the set of workflows in two sets, 
where we use 70\% of the workflows for training and the  remaining 30\% as the test workflows for the left-out dataset. 
That is the number of workflows in the train set is equal to $\lfloor 0.7 \times 35 \rfloor = 24$ which defines the 
number of hidden factors $r$. We use the 11 remaining workflows as  the   test set. For the both cold start scenarios, 
we report the average testing NDCG measure. We compute the average NDCG score at the truncation levels of $k=1,3,5$

For each of the two algorithms we presented we compute the number of times we have a performance win or loss compared to the performance 
of the baselines. On these win/loss pairs we do a McNemar's test of statistical significance and report its results, we set the significance level at $p=0.05$. 
For the matrix completion we denote the results of the performance comparisons against the CofiRank and LambdaMART baselines by
$\delta_{CR}$ and $\delta_{LM}$ respectively. We give the complete results in  table \ref{tab:mm_mc}.  
For the user cold start we denote the results of the performance comparisons against the user-memory based and the LambdaMART baselines
by $\delta_{UB}$ and $\delta_{LM}$ respectively, table~\ref{tab:mm_uc}. For the full cold start we denote by $\delta_{FB}$ and $\delta_{LM}$ 
the performance comparisons against the full-memory-based and the LambdaMART baselines respectively, table \ref{tab:mm_fc}. 
\note[Alexandros-June2014]{Fix your table references, for the two cold start matrices they are wrong.}

\subsubsection{Results}
\paragraph{Matrix Completion} CR achieves the lowest performance for $N=5$ and $N=10$, compared to the 
other methods; for $N=15$ the performances of all methods are very similar, table~\ref{tab:mm_mc}. The strongest
performance advantages over the CR baseline appear at $N=10$; there LMW and LM-MF are significantly better and LM-MF-Reg
is close to being significantly better, $p$-value=0.0824. At $N=5$ only LM-MF-Reg is significantly better
than CR. Overall it seems that using the side information in matrix completion problems brings a performance improvement
mainly when the preference matrix is rather sparse, as it is the case for $N=5$ an $N=10$, for lower sparsity
levels the use of side information does not seem to bring any performance improvement.
\note[Alexandros-June2014]{A pitty this is not explored in a more systematic way, along the discussions we had with the
guys. It might be one of the things for Florian to look in more detail.}

\paragraph{User Cold Start} LM-MF and LM-MF-Reg are the two methods that achieve the larger performance improvements over 
both the UB and the LM baselines for all values of the $k$ truncation level, table \ref{tab:mm_uc}. At $k=1$ both of them beat in a statistically
significant manner the UB baseline, and they are also close to beating in a statistical significant manner the LM baseline
as well, $p$-value=0.0636. At $k=3$ both beat in a statistically significant manner UB but not LM, at $=5$ there are no 
significant differences. Finally, note the LambdaMART variant that makes use of the weighted NDCG, LMW, does not seem 
to bring an improvement over the plain vanilla NDCG, LM. 

\paragraph{Full Cold Start} Here the performance advantage of LM-MF and LM-MF-Reg is even more pronounced, table \ref{tab:mm_fc}.
Both of them beat in a statistically significant manner the LM baseline for all values $k$ with a performance improvement of 0.7 in average. 
They also beat the FB baseline in a statistically significant manner at $k=1$ with LM-MF beating it also in a statistical significant manner at 
$k=3$ as well.

\note[Removed]{
The also beat in a statistically significant manner the FB baseline for $k=1,3$ with the little exception of the regularized variant which is close to be significant with $k=3$, $p$-value=0.1041. 
The performance drop compared to this baseline is now quite high where LM-MF-Reg has a drop of 0.13 for $k=1$ and LM-MF has a drop of 0.8 for $k=3$.  
However for $k=5$, FB achieves good performance thanks to the quality of the feature descriptors.   
At this level of $k$, it is close to beat in a significant manner LambdaMART and  significantly beats its weighted NDCG variant. 
Note that the meta-mining problem we have is a small dataset so it is rather difficult to beat the FB baseline. 
In addition remember that this recommendation task is the most difficult one among the three tasks. 
}

\subsubsection{Analysis}
\begin{figure}[!t] 
\centering
\subfloat[NDCG@1]{
	\includegraphics[width=2.5cm, height=2.5cm]{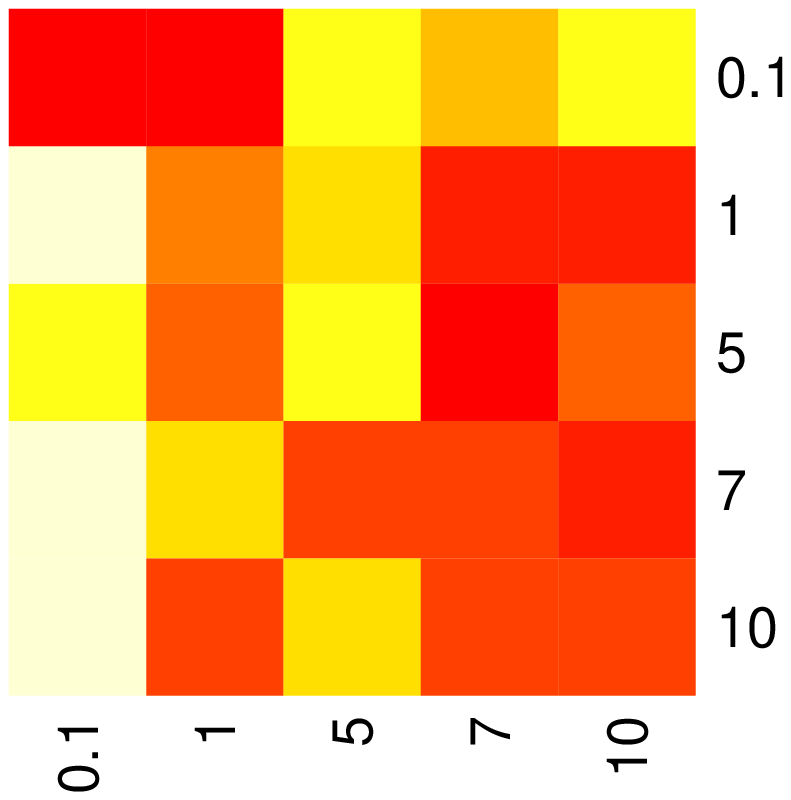}
}
\subfloat[NDCG@3]{
	\includegraphics[width=2.5cm, height=2.5cm]{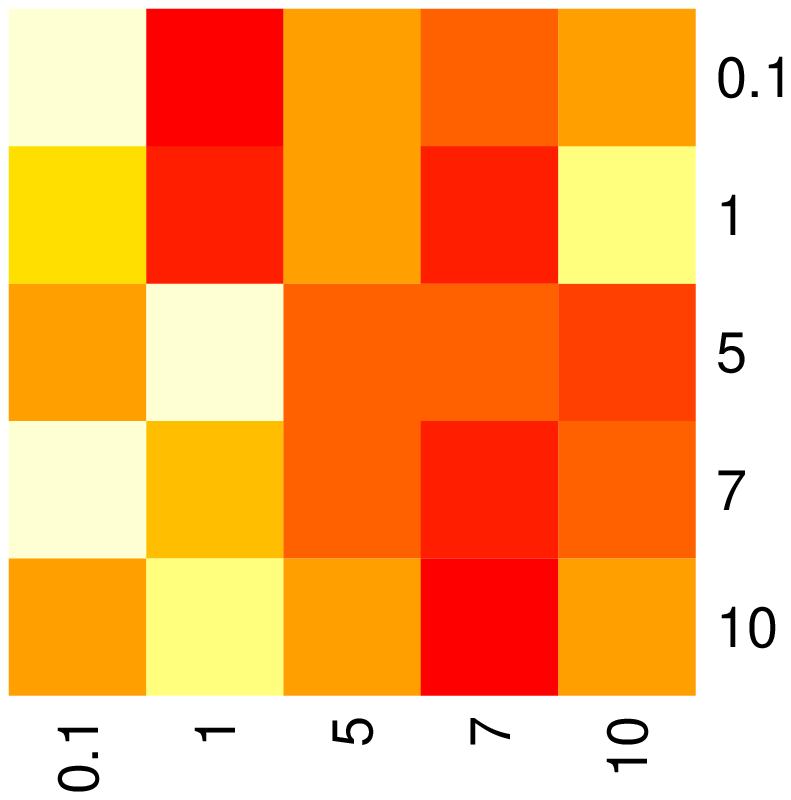}
}
\subfloat[NDCG@5]{
	\includegraphics[width=2.5cm, height=2.5cm]{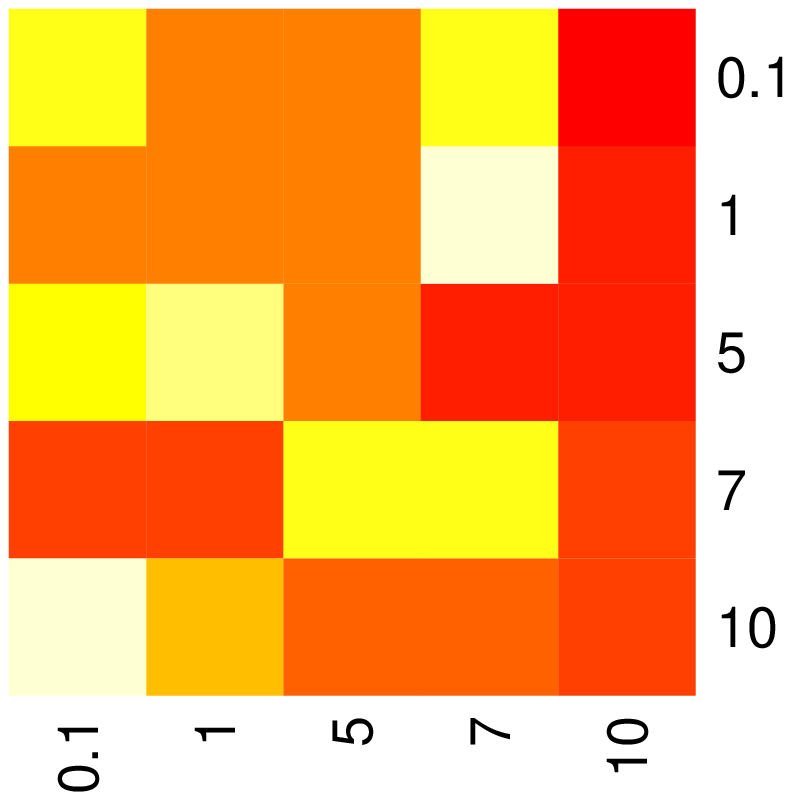}
}
\caption{$\mu_1$ and $\mu_2$ heatmap parameter distribution at the different truncation levels $k=1,3,5$ of NDCG@k from the {\em User Cold Start} meta-mining experiments. In the y-axis we have the $\mu_1$ parameter and in the x-axis the $\mu_2$ parameter. We validated each parameter with three-folds inner-cross validation to find the best value in the range $[0.1,1,5,7,10]$.}
\label{fig:pd}
\end{figure}

\begin{figure}[!t] 
\centering
\subfloat[NDCG@1]{
	\includegraphics[width=2.5cm, height=2.5cm]{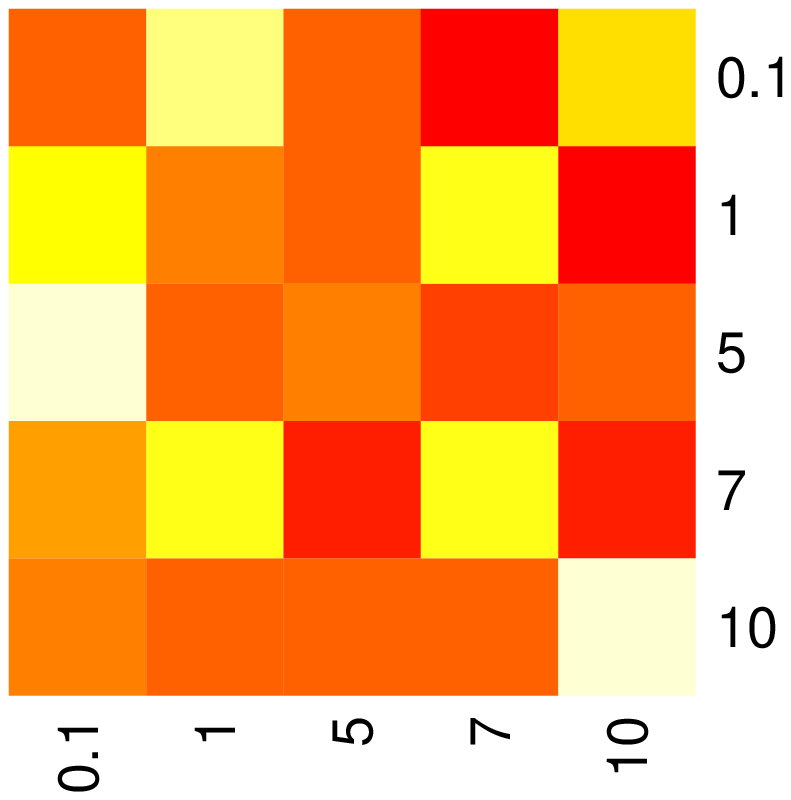}
}
\subfloat[NDCG@3]{
	\includegraphics[width=2.5cm, height=2.5cm]{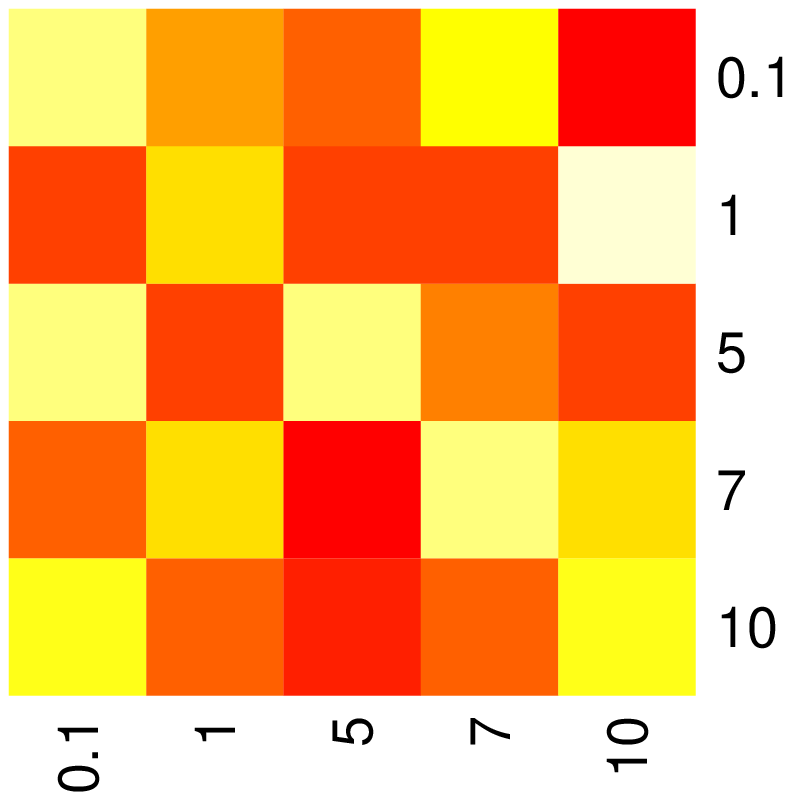}
}
\subfloat[NDCG@5]{
	\includegraphics[width=2.5cm, height=2.5cm]{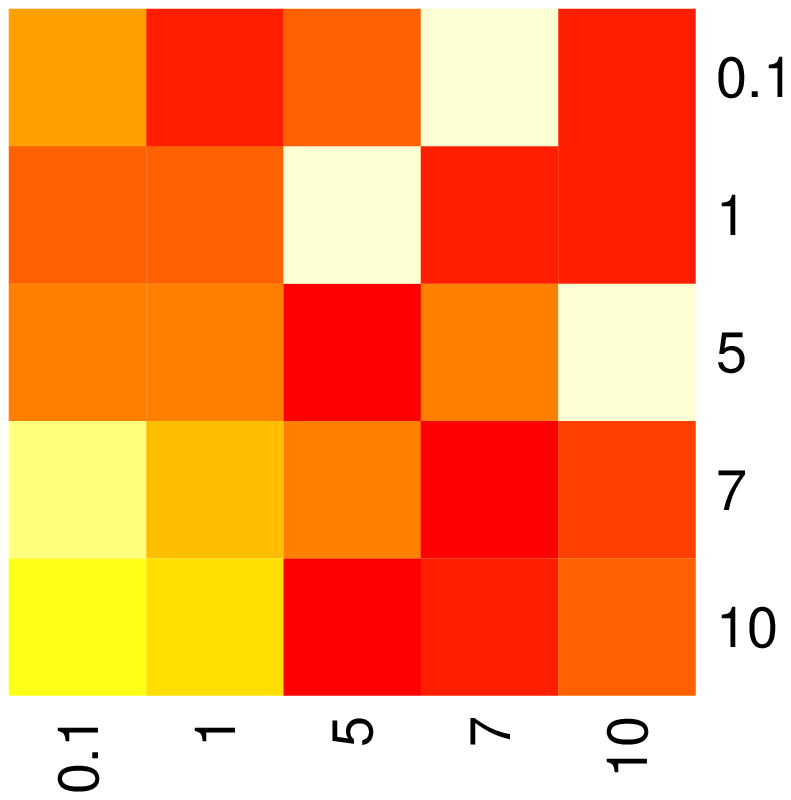}
}
\caption{$\mu_1$ and $\mu_2$ heatmap parameter distribution at the different truncation levels $k=1,3,5$ of NDCG@k from the {\em Full Cold Start} meta-mining experiments. The figure explanation is as before.}
\label{fig:pd2}
\end{figure}

\note[Removed]{From the results, the overall best method is the regularized variant, LM-MF-Reg. }
\note[Alexandros-June2014]{At least in terms of the significant wins, LM-MF has the advantage over LM-MF-Reg.}
  In order to study in more detail the effect of regularization we give in figures 
\ref{fig:pd} and \ref{fig:pd2} the frequency with which the two $\mu_1$ and $\mu_2$ 
input and output space  regularization parameters are set to the different possible values of the selection grid $[0.1, 1, 5, 7, 10]$,
over the 65 repetitions of the leave-one-dataset-out for the two cold start settings. 
The figures give the heatmaps of the pairwise selection frequencies; in the y-axis we have the $\mu_1$ parameter (input-space regularization hyperparameter) and in the 
x-axis the $\mu_2$ (output-space regularization hyperparameter).  A yellow to white cell corresponds to an upper mid to high selection frequency whereas an orange to red 
cell a bottom mid to low selection frequency. 

In the user cold start experiments, figure \ref{fig:pd}, we see that for $k=1$ the input space 
regularization is more important than the output space regularization, see the mostly white column 
in the (0.1, $\cdot$) cells. This makes sense since the latter regularizer uses only the 
top-one user or item preference to compute the pairwise similarities. As such, 
it does not carry much information to regularize appropriately the latent factors. 
For $k=3$ the  selection distribution spreads more evenlty over the different pairs 
of values which means that we have more combinations of the two regularizers. 
The most frequent pair is the (0.1, 0.1) which corresponds to low regularization, 
For $k=5$ we have a selection peak at (0.1, 10) and one at (7,1), it seems that
none of the two regularizers has a clear advantage over the other.

Looking now at the full cold start experiments, figure \ref{fig:pd2}, we have a rather different picture. 
For $k=1$ there is as before an advantage for the input space regularization where the (0.1, 5) cell is the most frequent one,  
however this time we also observe a high selection frequency on a strong input-output space regularization, the (10,10) selection peak.
This is also valid for $k=3$ where we have now a high selection frequency on strong regularization both for the input as well as the 
output space regulatization, see the selection peaks on, and around, (10,10). For $k=5$ we have now three peaks at (5, 1), (7, 0.1) 
and (10, 5) which indicate that the output space regularizer is more important than the input space regularizer. 

Overall in the full cold start setting  the regularization parameters seem to have more importance 
than in the user cold start setting.  This is reasonable since in the former scenario we have to predict  the 
preference of a new user-item pair, the latent profiles of which have to be regularized appropriately to avoid performance degradation.  
\note[Alexandros-June2014]{I do not understand what you try to say here, what do you mean by the sum of their different values is higher? 
In addition I am not sure there is a clear pattern emerging here, it seems that for the user cold start output space regularization 
does not have that much importance. For the Full cold start at k=1 and k=3 we need strong regularization on both input and output, the picture
changes at k=10 where only strong regularization on the output space seems to be required.}


\subsection{MovieLens}

The MovieLens recommendation dataset has a quite different morphology than that of the meta-mining. 
The descriptions of the users and the items are much more limited; users are described only by 
four features: sex, age, occupation and location, and movies by their genre which takes 19 different 
values, such as comedy, sci-fi to thriller, etc.  We model these descriptors using one-of-N representation. 

In the MovieLens dataset the side information has been mostly used to regularize the latent profiles and
not to predict them, see e.g \cite{Abernethy2006,Agarwal2010}. Solving the cold start problem in the presence
of so limited side information is a rather difficult task and typical collaborative filtering approaches make only use of the 
preference matrix $\mathbf Y$ to learn the user and items latent factors. 
\note[Alexandros-June2014]{What do you mean by rely only on the preference matrix?
Can you make it more clear?}

\subsubsection{Evaluation setting}
We use the same baseline methods as in the meta-mining problem.
We train LM with the following hyper-parameter setting: we fix the maximum number of nodes for the regression  
trees to 100, the maximum percentage of instances in each leaf node to $1\%$, and the learning rate, $\eta$, 
of the gradient boosted tree algorithm to $10^{-2}$.   As in the meta-mining problem for the construction of the 
ensemble of regression trees in LM-MF and LM-MF-Reg we use the same setting for the hyperparameters they share with 
LM. We optimize their $\mu_1$ and $\mu_2$ regularization
parameters within the grid  $[0.1, 1, 5, $ $10, 15, 20, 25, 50, 80, 100]^2$ using five-fold inner cross validation. 
We fix the number of factors $r$ to 50 and the number of nearest neighbors in the Laplacian matrices to five. 
As in the meta-mining problem we used the NDCG similarity measure to define output-space similarities with the 
truncation level $k$ set to the same value as the one at which we evaluate the methods.

We will evaluate and compare the different methods under the two cold start scenarios we described previously. 
In the user cold start problem, we randomly select 50\% of the users for training and we test the 
recommendations that the learned models generate on the remaining 50\%. 
In the full cold start problem, we use the same separation to training and testing users and  
in addition we also randomly divide the item set to two equal size subsets, where we use one
for training and the other for testing. Thus the learned models are evaluated on users and 
items that have never been seen in the model training phase. 
We give the results for the 100k and 1M variants of the MovieLens dataset 
for the user cold start scenario in table~\ref{tab:ml1} and for the full cold start scenario in table~\ref{tab:ml2}.
\note[Alexandros-June2014]{Fix the references to the tables, they do not appear correctly.}

\subsubsection{Results}
In the user cold start scenario, table~\ref{tab:ml1}, we can see first that both LM-MF and LM-MF-Reg 
beat in a statistically significant manner the UB baseline in both MovieLens variants for all values 
of the truncation parameter, with the single exception of LM-MF at 100k and $k=5$ for which the performance difference is not statistically significant. 
LM and its weighted NDCG variant, LMW,  are never able to beat UB; LM is even statistically significantly worse compared to UB at 1M for $k=5$. 
Moreover both LM-MF and its regularized variant beat in a statistically significant manner LM in both MovieLens variants and for all values of $k$. 
LM-MF-Reg has a small advantage over LM-MF for 100K, it achieves a higher average NDCG score, however this advantage disappears
when we move to the 1M dataset, which is rather normal since due to the much larger dataset size 
there is more need for additional regularization.  Moreover since we do a low-rank matrix factorization, we have $r=50$ for the 
roughly $6000$ users. \note[Alexandros-June2014]{roughly 6000 what? users? items?}
which already is on its own a quite strong regularization rendering also unnecessary 
the need for the input/output space regularizers. 
 
A similar picture arises in the full cold start scenario, table~\ref{tab:ml2}. 
With the exception of MovieLens 100K at $k=10$ both LM-MF and LM-MF-Reg beat always in a statistically significant the FB baseline. 
Note also that now LM as well as its weigthed NDCG variant are significantly better than the FB baseline for the 1M dataset and $k=5,10$.
In adition the  weigthed NDCG variant is significantly better than LM for 100k and 1M at $k=5$ and $k=10$ respectively.
Both  LM-MF and LM-MF-Reg beat in a statistically significant manner LM, with the exception of 100k at $k=10$ where there is 
no significant difference. Unlike the user cold start problem, now it is LM-MF that achieves the best overall performance.

\note[Removed]{
There is no clear reason why we get this result for the full cold start scenario but this can be explained by the low 
description of movies in MovieLens which are only described by their genre, resulting in movie input/output space regularizers which are 
not as useful as the user regularizers in the user cold start setting.
} 
\note[Alexandros-June2014]{You say that there is no reason
for that result and then you provide an explanation... So is there a reason or not? You seem to be saying the movie features 
are not that good, but one could have said the same thing for the user features, which make difficult to explain the better
performance of LM-MF-reg at the user cold start scenarion.} 

\begin{table*}[!ht]
\centering
\begin{tabular}{|l|l|l|l|}
\hline
 
 &  $N=5$ &  $N=10$ &  $N=15$ \\ \hline
CR & 0.5755 & 0.6095 & 0.7391 \\ 
\hline
LM & 0.6093 & 0.6545 & 0.7448 \\ 
$\delta_{CR}$ & p=0.4567(=) & p=0.0086(+) & p=1(=) \\ 
\hline
LMW & 0.5909 & 0.6593 & 0.7373 \\ 
$\delta_{CR}$ & p=1(=) & p=0.0131(+)  & p=0.6985(=) \\ 
$\delta_{LM}$ & p=0.2626 & p=0.0736(=)  & p=1(=) \\ 
\hline
LM-MF & 0.6100 & {\bf 0.6556} & 0.7347 \\ 
$\delta_{CR}$ & p=0.0824(=) & p=0.0255(+)  & p=0.6985(=) \\ 
$\delta_{LM}$ & p=0.7032(=) & p=1(=) & p=0.7750(=) \\ 
\hline
LM-MF-Reg & {\bf 0.61723} & 0.6473 & {\bf 0.7458} \\ 
$\delta_{CR}$ & p=0.0471(+) & p=0.0824(=)  & p=0.6985(=) \\ 
$\delta_{LM}$ & p=0.9005(=) & p=0.8955(=) & p=0.2299(=) \\ 
\hline
\end{tabular}
\caption{
NDCG@5 results on meta-mining for the {\em Matrix Completion} setting. 
$N$ is the number of workflows we keep in each dataset for training. 
For each method, we give the comparison results against the CofiRank and LambdaMart methods in the rows denoted by $\delta_{CR}$ and $\delta_{LM}$ respectively. 
More precisely we report the $p$-values of the McNemar's test on the numbers of wins/losses and denote by (+) a statistically significant improvement, by (=) no performance difference and by (-) a significant loss. In bold, the best method for a given $N$. 
}
\label{tab:mm_mc}
\end{table*}

\begin{table*}[!ht]
\centering
\subfloat[{\em User Cold Start}]{
\label{tab:mm_uc}
\begin{tabular}{|l|l|l|l|}
\hline
	& k=1 & k=3 & k=5 \\ \hline
UB & 0.4367&  0.4818 & 0.5109  \\ \hline
LM & 0.5219 & 0.5068 & 0.5135  \\ 
$\delta_{UB}$ & p=0.0055(+)  & p=0.1691(=)  & p=0.9005(=)    \\ \hline
LMW & 0.5008  & 0.5232  & 0.5168  \\ 
$\delta_{UB}$ &  p=0.0233(+) & p=0.2605(=)  & p=0.5319(=)   \\ 
$\delta_{LM}$ & p=0.4291(=)  & p=0.1845(=) & p=0.8773(=)   \\ 
\hline
LM-MF & 0.5532 & {\bf 0.5612} & {\bf 0.5691} \\ 
$\delta_{UB}$ & p=0.0055(+)  & p=0.0086(+) &  p=0.3210(=)    \\
$\delta_{LM}$ & p=0.0636(=)  &  p=0.0714(=)  & p=0.1271(=)   \\ 
\hline
LM-MF-Reg & {\bf 0.5577} & 0.5463 & 0.5569 \\ 
$\delta_{UB}$ & p=0.0055(+)  & p=0.0086(+) & p=0.3815(=)    \\
$\delta_{LM}$ & p=0.0636(=)  & p=0.1056(=)  & p=0.2206(=)   \\ 
\hline
\end{tabular}
}
\subfloat[{\em Full Cold Start}]{
\label{tab:mm_fc}
\begin{tabular}{|l|l|l|l|}
\hline
 & $k=1$ & $k=3$ & $k=5$ \\ \hline
FB & 0.46013 & 0.5329 & 0.5797 \\ \hline
LM & 0.5192 & 0.5231 & 0.5206 \\ 
$\delta_{FB}$ & p=0.0335(+) & p=0.9005(=) & p=0.0607(=) \\ 
\hline
LMW & 0.5294 & 0.5190  & 0.5168  \\ 
$\delta_{FB}$ &  p=0.0086(+)  &  p=1(=) & p=0.0175(-)   \\ 
$\delta_{LM}$ &  p=1(=) & p=1(=)  & p=0.5218(=)   \\ 
\hline
LM-MF & 0.5554 & {\bf 0.6156} & 0.5606 \\ 
$\delta_{FB}$ & p=0.0175(+) & p=0.0175(+) & p=0.6142(=) \\ 
$\delta_{LM}$ & p=0.0171(+) & p=0.0048(+) & p=0.0211(+) \\ 
\hline
LM-MF-Reg & {\bf 0.5936} & 0.5801 & {\bf 0.5855} \\ 
$\delta_{FB}$ &  p=0.0007(+) & p=0.1041(=) & p=1(=) \\ 
$\delta_{LM}$ &  p=0.0117(+) & p=0.0211(+) &  p=0.0006(+) \\ 
\hline 
\end{tabular}
}
\caption{
NDCG@k results on meta-mining for the two cold start settings. 
For each method, we give the comparison results against the  user, respectively full, memory-based and LambdaMart methods in the rows denoted by $\delta_{UB}$, respectively $\delta_{FB}$, and $\delta_{LM}$. 
The table explanation is as before. In bold, the best method for a given $k$.
}
\end{table*}

\begin{table*}[t]
\centering
\scalebox{0.8}{
\subfloat[{\em User Cold Start}]{
\label{tab:ml1}
\begin{tabular}{|l|l|l|l|l|}
\hline
 & \multicolumn{2}{|c|}{100K} & \multicolumn{2}{|c|}{1M} \\ \hline
 & $k=5$ & $k=10$ & $k=5$ & $k=10$ \\ \hline
UB & 0.6001 &  0.6159 & 0.6330  & 0.6370  \\ \hline
 LM & 0.6227 & 0.6241 & 0.6092 & 0.6453 \\ 
 $\delta_{UB}$ & p=0.2887(=)  & p=0.8537(=) & p=0.0001(-) & p=0.1606(=) \\ 
\hline
 LMW & 0.6252  & 0.6241  & 0.6455 & 0.6450 \\ 
 $\delta_{UB}$ & p=0.0878(=)  & p=0.5188(=) & p=0.0120(+) & p=0.1448(=) \\ 
 $\delta_{LM}$ & p=0.7345(=) & p=0.9029(=) & p=0.0000(+) & p=0.5219(=) \\ 
\hline
  LM-MF & 0.6439 &  0.6455 & {\bf 0.6694} & {\bf 0.6700} \\ 
 $\delta_{UB}$ & p=0.0036(+) & p=0.1171(=) & p=0.0000(+) & p=0.0000(+) \\ 
 $\delta_{LM}$ & p=$2e^{-06}$(+) & p=$6e^{-06}$(+) & p=0.0000(+) & p=0.0000(+) \\ 
\hline
   LM-MF-Reg & {\bf 0.6503} & {\bf 0.6581} & {\bf 0.6694} & {\bf 0.6705} \\ 
 $\delta_{UB}$ & p=$4e^{-05}$(+) & p=0.0001(+) & p=0.0000(+) & p=0.0000(+) \\ 
 $\delta_{LM}$ & p=$3e^{-06}$(+) & p=0.0000(+) & p=0.0000(+) & p=0.0000(+) \\ 
\hline
\end{tabular}
}
\subfloat[{\em Full Cold Start}]{
\label{tab:ml2}
\begin{tabular}{|l|l|l|l|l|}
\hline
 &  \multicolumn{2}{|c|}{100K} &  \multicolumn{2}{|c|}{1M} \\ \hline
 &  $k=5$ & $k=10$ &  $k=5$ & $k=10$ \\ \hline 
FB & 0.5452 & 0.5723 & 0.5339 & 0.5262 \\ 
\hline
LM & 0.5486 & 0.5641& 0.5588 & 0.5597 \\ 
$\delta_{FB}$ &  p=0.7817(=) & p=0.4609(=) & p=$1e^{-06}$(+) & p=0.0001(+)  \\ 
\hline
LMW & 0.5549  & 0.5622  & 0.55737 & 0.5631  \\ 
 $\delta_{FB}$ &  p=0.4058(=)  &  p=0.2129(=) & p=$9e^{-06}$(+) &  p=$1e^{-06}$(+) \\ 
 $\delta_{LM}$ &  p=0.0087(+) & p=0.8830(=) & p=0.1796(=) &  p=$3e^{-06}$(+) \\ 
\hline
LM-MF & {\bf 0.5893} & {\bf 0.5876} & {\bf 0.5733} & {\bf 0.5750} \\ 
$\delta_{FB}$ &  p=0.0048(+) &  p=0.2887(=) & p=0.0000(+) &  p=0.0000(+) \\ 
$\delta_{LM}$ &  p=$3e^{-06}$(+) &  p=0.0001(+) &  p=0.0000(+) &  p=0.0000(+) \\ 
\hline
LM-MF-Reg & 0.5699 & 0.57865 & {\bf 0.5736} & 0.5683 \\ 
$\delta_{FB}$ & p=0.0142(+) & p=0.1810(=) &  p=0.0000(+) &  p=$4e^{-07}$(+) \\ 
$\delta_{LM}$ &  p=0.0310(+) & p=0.0722(=) &  p=0.0000(+) &  p=0.0000(+) \\ 
\hline
\end{tabular}
}
}
\caption{
NDCG@k results on the two MovieLens datasets for the two cold start setting. 
For each method, we give the comparison results against the user, respectively full, memory-based and LambdaMART methods in the rows denoted by $\delta_{UB}$, respectively $\delta_{FB}$, and $\delta_{LM}$. 
The table explanation is as before. 
In bold, the best method for a given $k$.
}
\end{table*}

\section{Conclusion}
\label{sec:conc}

Since only top items are observable by users in real recommendation systems, we believe that ranking loss functions that focus on the correctness 
of the top item predictions are more appropriate for this kind of problem. We explore the use of a state of the art learning to rank algorithm, 
LambdaMART, in a recommendation setting with an emphasis on the cold-start problem one of the most challenging problems in recommender systems. 
However plain vanilla LambdaMART has a certain number of limitations which spring namely from the fact 
that it lacks a principled way to control overfitting relying in ad-hoc approaches. We proposed a number of ways to deal with these limitations. 
The most important is that we cast the learning to rank problem as learning a low-rank matrix factorization; our underlying
assumption here being that the descriptions of the users and items as well as their preference behavior are governed by a few latent factors.
The user item-preferences are now computed as inner products of the learned latent representations of users and items. In addition to the regularization effect of the low rank factorization 
we bring in additional regularizers to control the complexity of the latent representations, regularizers which reflect the users and items 
manifolds as these are explicited by user and item feature descriptions as well as the preference behavior.  We report results on two very 
different recommendation problems, meta-mining and MovieLens, and show that the performance of the algorithms we propose beats in a statistically
significant manner a number of baselines often used in recommendation systems.



\bibliographystyle{abbrv}
\bibliography{ref}  

\end{document}